%% file: main.tex
\documentclass{article}

\usepackage{microtype}
\usepackage{graphicx}
\usepackage{subfigure}
\usepackage{booktabs} 

\usepackage{hyperref}

\usepackage[accepted]{icml2024}

\usepackage{amsmath}
\usepackage{amssymb}
\usepackage{mathtools}
\usepackage{amsthm}
\usepackage{textgreek}

\usepackage[capitalize,noabbrev]{cleveref}

\theoremstyle{plain}

\theoremstyle{definition}

\theoremstyle{remark}

\usepackage[textsize=tiny]{todonotes}

\newcommand{\lgreenb}[1]{\colorbox[HTML]{ABDCCC}{#1}}
\newcommand{\greenb}[1]{\colorbox[HTML]{02D89E}{#1}}
\newcommand{\lredb}[1]{\colorbox[HTML]{F5A49D}{#1}}
\newcommand{\redb}[1]{\colorbox[HTML]{F56055}{#1}}
\newcommand{\greyb}[1]{\colorbox[HTML]{D6D6D6}{#1}}

\definecolor{tred}{rgb}{0.569, 0.172, 0.243}
\newcommand{\tred}[1]{\textcolor{tred}{#1}}
\definecolor{tblue}{rgb}{0.236, 0.439, 0.599}
\newcommand{\tblue}[1]{\textcolor{tblue}{#1}}

\definecolor{fblue}{rgb}{0.294, 0.462, 0.666}
\newcommand{\fblue}[1]{\textcolor{fblue}{#1}}
\definecolor{forange}{rgb}{0.874, 0.501, 0.286}
\newcommand{\forange}[1]{\textcolor{forange}{#1}}
\definecolor{fred}{rgb}{0.717, 0.278, 0.227}
\newcommand{\fred}[1]{\textcolor{fred}{#1}}
\definecolor{fgreen}{rgb}{0.4, 0.619, 0.298}
\newcommand{\fgreen}[1]{\textcolor{fgreen}{#1}}
\definecolor{fpurple}{rgb}{0.537, 0.423, 0.705}
\newcommand{\fpurple}[1]{\textcolor{fpurple}{#1}}
\definecolor{fbrown}{rgb}{0.501, 0.360, 0.321}
\newcommand{\fbrown}[1]{\textcolor{fbrown}{#1}}
\definecolor{fcyan}{rgb}{0.788, 0.513, 0.729}
\newcommand{\fcyan}[1]{\textcolor{fcyan}{#1}}
\definecolor{fgrey}{rgb}{0.505, 0.505, 0.501}
\newcommand{\fgrey}[1]{\textcolor{fgrey}{#1}}
\usepackage{inconsolata}

\icmltitlerunning{Data Engineering for Scaling Language Models to 128K Context}

\begin{document}

\twocolumn[
\icmltitle{Data Engineering for Scaling Language Models to 128K Context}



\icmlsetsymbol{equal}{*}




\begin{center}
\textbf{Yao Fu$^\text{\textkappa}$ \quad Rameswar Panda$^\text{\texteta}$ \quad Xinyao Niu$^\text{\textmu}$\quad Xiang Yue$^\text{\textpi}$ \quad 
Hannaneh Hajishirzi$^\text{\textsigma}$\quad Yoon Kim$^\text{\textlambda}$ \quad Hao Peng$^\text{\textdelta}$
}\\
$^\text{\textkappa}$University of Edinburgh\quad $^\text{\texteta}$MIT-IBM Watson AI Lab\quad $^\text{\textmu}$University of Melbourne\quad $^\text{\textpi}$ Ohio State University \\
$^\text{\textsigma}$University of Washington\quad  $^\text{\textlambda}$MIT\quad $^\text{\textdelta}$UIUC \\
\texttt{yao.fu@ed.ac.uk}\quad\texttt{yoonkim@mit.edu}\quad\texttt{haopeng@illinois.edu}\\
\url{https://github.com/FranxYao/Long-Context-Data-Engineering}
\end{center}

\icmlkeywords{Machine Learning, ICML}

\vskip 0.3in
]




\begin{abstract}
We study the continual pretraining recipe for scaling language models' context lengths to 128K, with a focus on data engineering.
We hypothesize that  long context modeling, in particular \textit{the ability to utilize information at arbitrary input locations}, is a capability that is  
mostly already acquired through large-scale pretraining, and that this capability can be readily extended to contexts substantially longer than seen during training~(e.g., 4K to 128K) through lightweight continual pretraining on appropriate data mixture. 
We investigate the \textit{quantity} and \textit{quality} of the data for continual pretraining: 
(1) for quantity, we show that 500 million to 5 billion tokens are enough to enable the model to retrieve information anywhere within the 128K context;
(2) for quality, our results equally emphasize \textit{domain balance} and \textit{length upsampling}.
Concretely, we find that na\"{i}vely upsampling longer data on certain domains like books, a common practice of existing work, gives suboptimal performance, and that a balanced domain mixture is  important.
We demonstrate that continual pretraining of the full model on 1B-5B tokens of such data is an effective and affordable strategy for scaling the context length of language models to 128K.
Our recipe outperforms strong open-source long-context models
and closes the gap to frontier models like GPT-4 128K.
\end{abstract}

\section{Introduction}
\label{sec:intro}
\input{010_intro}

\section{Background}
\label{sec:background}
\input{020_background}

\section{Long Context Data Composition}
\label{sec:data}
\input{030_data_composition}

\section{Infrastructure and Engineering}
\label{sec:exp_setting}
\input{040_exp_setting}

\section{Experimental Results}
\label{sec:exp}
\input{050_exp}

\section{Discussion}

\label{sec:discussions}
\input{060_discussion}

\section{Conclusion}

\label{sec:conclusion}
\input{070_conclusion}



\bibliography{long_context}
\bibliographystyle{icml2024}

\newpage
\appendix
\onecolumn



\end{document}

%% file: 010_intro.tex
\input{figures/fig_needle_in_haystack}

A context window of 128K tokens enables large language models to perform tasks that significantly beyond existing paradigm, such as multi-document question answering~\citep{caciularu2023peek}, repository-level code understanding~\citep{bairi2023codeplan}, long-history dialog modeling~\citep{mazumder2024lifelong}, and language model-powered autonomous agents~\citep{weng2023prompt}.  
A popular testbed for whether models can actually utilize long context length is the recent Needle-in-a-Haystack test~\citep{needleinhaystack}, which asks the model to precisely recite the information in a given sentence where the sentence (the ``needle'') is placed in an arbitrary location of a 128K long document~(the ``haystack''). 
In the open-source space, although works like LongLoRA~\citep{chen2023longlora} and YaRN-Mistral~\citep{peng2023yarn} theoretically support 100K context, they are not able to pass this test at such context lengths, as shown in Fig.~\ref{fig:needle_in_a_haystack}. 
Currently, only closed-source frontier models like GPT-4 128K have demonstrated strong performance on the Needle-in-a-Haystack test.

This work investigates data engineering methods for scaling language models' context lengths. 
Our objective is to continue pretraining the language model on appropriate data mixtures  such that it can  pass the Needle-in-a-Haystack test at  128K length. 
Given that most  existing models are trained on less than 4K context length~\citep{touvron2023llama} and that attention has quadratic complexity, continual pretraining with full attention on much longer context lengths (we train on 64K-80K context lengths) may seem prohibitively costly at a first glance. However, we show that this is {feasible under academic-level resources} (see Table~\ref{tab:infra}).
We use LLaMA-2 7B and 13B as our base models. 
We do not make any significant change to model architecture other than adjusting the base of RoPE, as in~\citet{xiong2023effective}.
Our major focus is the data recipe: \textit{what} and \textit{how much} data is able to well-adapt a model to pass the Needle-in-a-Haystack test at 128K context length.

We hypothesize that the capability to utilize
information at arbitrary locations within long
context length is (mostly) already acquired during pretraining, even for
models pretrained on substantially shorter
4K contexts. 
This hypothesis is in contrast to existing works like~\citet{xiong2023effective, xverse}, which perform continual pretraining on a large amount of data (400B tokens) to \textit{inject} long-context-modeling capabilities; in this strategy, the cost can be as high as pre-training from scratch. 
In this work we show that continual pretraining on a small amount of long-context data, in our case, 1-5B tokens, can ``unlock'' a 7B model's capability of precise retrieval over much longer context lengths than seen in original pretraining.

We further show that upsampling long sequences while retaining the domain mixture of the pretraining corpora is crucial for context scaling, yet overlooked by existing works (Table~\ref{tab:design_diff}).
Most existing works are based on the following intuition: to model long-range dependencies one needs long sequence data, which domains like books provide; therefore, it is necessary to upsample domains containing long sequences in the data mixture, as done by LongChat 32K~\citep{longchat2023} and YaRN Mistral 128K~\citep{peng2023yarn}. 
However, we show that this intuitive solution is suboptimal because, as we observe, this results in perplexiy degradations in other domains (Table~\ref{tab:loss_7b}).
Instead, a data mixture that keeps the domain mixture ratio the same as the pretraining mixture, and then upsampling long sequences within each domain gives the most stable performance gain.
We give evidence that this is the primary reason our solution improves long context tasks while maintaining short context performance, compared with strong baselines like YaRN-Mistral 128K~\citep{peng2023yarn} and LongLoRA 100K~\citep{chen2023longlora}.

In summary, we propose a concrete data recipe for scaling language model context length to 128K,
specifically, which involves {continual pretrain the full-attention model on 1-5B tokens of per-source-length upsampled data}.
We show that our recipe results in 7B and 13B LLaMA-2 of strong long-context performance, substantially closing the gap to frontier models like GPT-4 128K on the Needle-in-a-Haystack test,
opening future possibilities of studying long-context modeling under academic budgets.

%% file: figures/fig_needle_in_haystack.tex
\begin{figure*}[!t]
\small
  \centering
  \includegraphics[width=\linewidth]{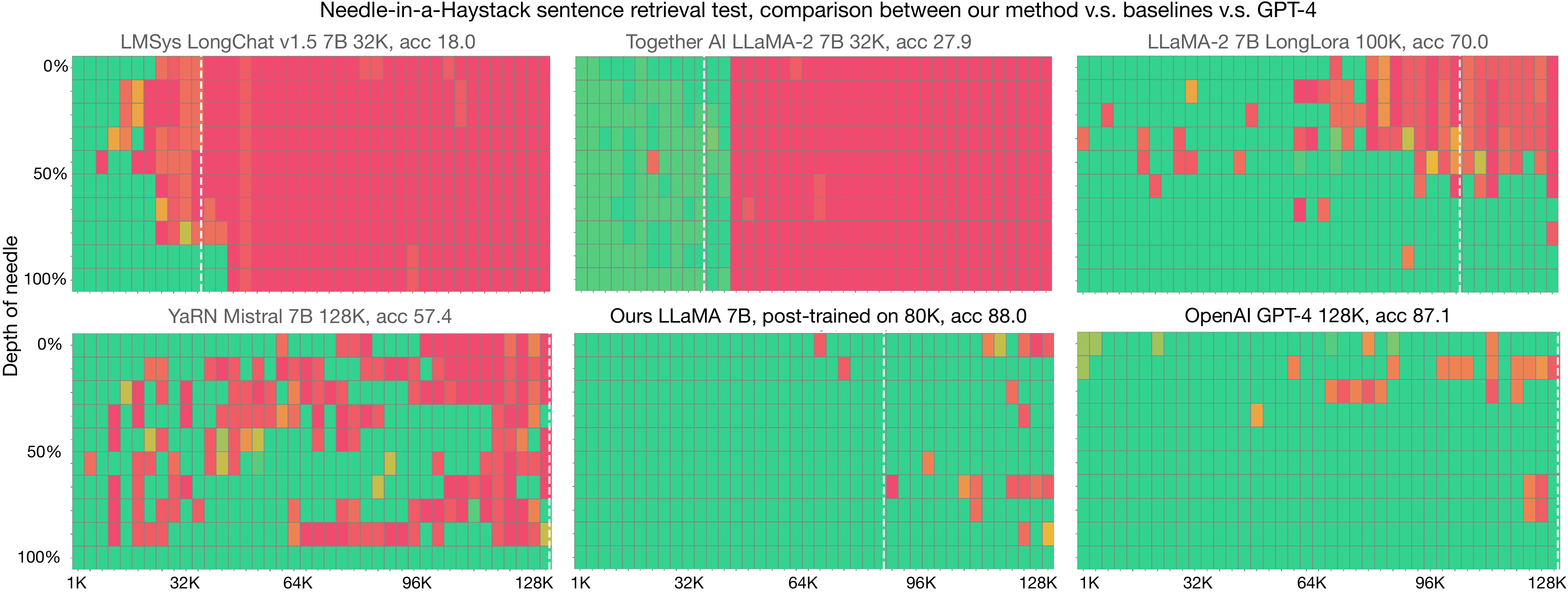}
    \vspace{-6mm}
  \caption{
    Needle-in-a-Haystack test~\citep{needleinhaystack} performance comparison.
    The x-axis denotes the length of the document (the ``haystack''); the y-axis indicates the position that the ``needle'' (a short sentence) is located within the document, from 1K to 128K tokens. For example, 50\% indicates that the needle is placed in the middle of the document.
    A red cell means the the language model cannot recite the information in the needle, and a green cell means the model can. A white dashed line means models' continual pretraining (or finetuning for instruction-tuned models) context length; thus the area to its right indicates length generalization. Most  existing open-source models make mistakes when the document is long.
    Our post-training recipe demonstrates strong performance up to about 100K  length.
  }
    \vspace{-2mm}
  \label{fig:needle_in_a_haystack}
\end{figure*}

%% file: 020_background.tex
\input{tables/tab_design_diff}

Frontier language models feature extremely long context lengths, such as OpenAI's GPT-4 Turbo 128K (Nov 2023) and Anthropic's Claude-100K (May 2023). 
In the regime of 100K, a wide array of new applications emerge, such as repo-level code understanding~\citep{bairi2023codeplan}, long-history dialog modeling~\citep{mazumder2024lifelong}, and language model-powered autonomous agents~\citep{weng2023prompt}.
A recent testbed for long-range capabilities is the Needle-in-a-Haystack benchmark, first proposed by~\citet{needleinhaystack} in Nov 2023. This benchmark asks the language model to recite the information in a ``needle'' sentence (``The best thing to do in San Francisco is eat a sandwich and sit in Dolores Park on a sunny day'') that is randomly inserted at an arbitrary location in a long essay.
Since its release, it has become a popular sandbox for testing whether models can utilize 100K+ context lengths, as it differentiates models that can precisely recite the given information at arbitrary input location versus those models that cannot.
So far there is neither public knowledge nor open-source work about achieving precise retrieval anywhere at this scale to the best of our knowledge. 

This work improves the long-context capability over strong open-source baselines, and closes the gap to GPT-4 on the Needle-in-a-Haystack benchmark, as demonstrated in Fig.~\ref{fig:needle_in_a_haystack}.
Our baselines here include LMSys' LongChat v1.5 32K~\citep{longchat2023}, Together Compute's LLaMA-2 32K~\citep{togetherllama32k}, YaRN Mistral 128K~\citep{peng2023yarn}, and LongLoRA~\citep{chen2023longlora}, which are so  far the top open-source long-context language models. 
These works focus on different aspects of long-context language modeling. For example, YaRN~\citep{peng2023yarn} focuses on positional embeddings, LongLoRA~\citep{chen2023longlora} focuses on efficient attention, and Together~\citep{togetherllama32k} focuses on a full-stack solution. 
Our work focuses on data-engineering, and identifies critical data aspects for extending language models' long-term dependency modeling to much longer contexts than seen during regular pretraining. 

The major differences between this work and existing work are listed on Table~\ref{tab:design_diff}.
Together's LLaMA-2 is trained on 32K, but only generalizes to about 40K length. 
YaRN Mistral is trained on book-only data; later in Table~\ref{tab:loss_7b} we will show that improvements on one domain has limited transfer to other domains, indicating that one should consider a balanced mixture of different domains.
LongLoRA does not upsample long sequences; later in Fig.~\ref{fig:original_vs_per_source} we will show that upsampling long sequences is critical for precise retrieval. 
These details are relatively hard to notice, yet we believe that they are the important for improved performance. 
Before the Needle-in-a-Haystack test, most  existing works use test negative log-likelihood as evaluation, 
effectively {concealing} the underlying differences beyond low loss.
We  show that, similar test loss could result in substantially different behavior when performing precise retrieval (Fig.~\ref{fig:original_vs_per_source}). 

Another important related work is the previous LLaMA Long~\citep{xiong2023effective} work and the concurrent XVERSE~\citep{xverse} work, which continue pretraining the model on 32K sequences for about {500 billion tokens}. 
These works are implicitly motivated by the view that long-context modeling is a new capability that must be ``injected'' through large-scale training.
We instead hypothesize that the base model has mostly already acquired   this capability through large-scale pretraining, and thus a lightweight continual pretraining on relatively small data (e.g., 5B tokens) is enough to extend these capabilities to much longer context lengths  (Fig.~\ref{fig:needle_data_scaling}).

%% file: tables/tab_design_diff.tex
\begin{table*}[t!]
\centering
\caption{
Major differences between \tblue{our method} v.s. \tred{existing work} from a data perspective, which we view as critical for extending context length.
 Our data mixture differs from previous work in  three ways: 
(1) \textit{length of continual pretraining data}: we use 80K compared to Together's 32K, which does not generalizes beyond 32K;
(2) \textit{data mixture}: we use SlimPajama which has balanced domains compared to YaRN, which uses book-only PG19;
(3) \textit{length upsampling}: we upsample long sequences compared to LongLoRA, which does not.
Despite the subtlety of these details (e.g., many of these details are just mentioned as a single line in previous works), we find that these details are crucial for performance on long-range retrieval.
}
\begin{tabular}{@{}lll@{}}
\toprule
                   &\bf Existing data solution &\bf Our solution  \\ 
\midrule  
Together LLaMA-2 32K    & Train on \tred{\underline{32K}} length  & Train on \tblue{\underline{80K}} length   \\ 
LongChat v1.5 32K & Train on \tred{\underline{32K}} length \tred{\underline{dialog data}} & Train on \tblue{\underline{80K}} length-upsampled \tblue{\underline{SlimPajama}} \\ 
YaRN Mistral 7B 128K    & Train on \tred{\underline{PG19 book-only}}  & Trained on length-upsampled \tblue{\underline{SlimPajama}} \\ 
LongLoRA  100K          & Train on Redpajama \tred{\underline{no length upsampling}}  & With \tblue{\underline{length upsampling}}    \\ 
\bottomrule
\end{tabular}
\label{tab:design_diff}
\end{table*}

%% file: 030_data_composition.tex
\input{figures/fig_data_mix}

We use the SlimPajama~\citep{cerebras2023slimpajama} dataset for continual pretraining. 
This dataset is an open-source reproduction of the LLaMA~\citep{touvron2023llama} pretraining data mixture, 
consisting of 82\% web data (67\% from CommonCrawl and 15\% from C4), 4.5\% code (Github), 4.5\% Wikipedia, 4.5\% books, 2.5\% Arxiv, and 2.0\% StackExchange. 
Since this dataset closely mirrors that used to pretrain the LLaMA models,
there is less concern of distribution shift during continual pretraining; it is therefore used by many recent works like~\citet{openmoe2023}. 

The documents' lengths and their source domains are two closely related confounding factors in data engineering because long data usually come from  particular sources.
Our examination shows books and Github code are the longest sources, followed by Arxiv. 
Webpages like C4 and StackExchange tend to be shorter. 
Directly upsampling long data changes the domain mixture, e.g., upsampling sequences longer than 100K will increase the portion of the books domain.
Likewise, changes in the domain mixture will result in shifts of the length distribution. 
Our guideline is to decompose these two factors step by step, as is shown in Fig.~\ref{fig:data_mix}.
We consider the following methods: 

\textbf{Cut at 4K:}\quad This  truncates documents into 4K-length chunks. This approach inherits the original data mixture and is used by most pretraining works, such as~\citep{touvron2023llama, hoffmann2022training}. Since there are about 30\% documents that are naturally longer than 4K, this approach breaks such naturally-existing long-range dependencies.

\textbf{Cut at 128K:}\quad This preserves most of the naturally-existing long-range dependencies without changing the domain mixture. LongLoRA~\citep{chen2023longlora} follows this approach. However, we will show later that only using the naturally-existing long dependencies is inadequate  (Fig.~\ref{fig:original_vs_per_source}).

\textbf{Per-source Upsampling:}\quad This retains the domain mixture, then upsamples long documents within each domain. This approach upsamples long documents without changing the domain mixture. We recommend this approach as our experiments suggest that this gives the most balanced performance gain.

\textbf{Global Upsampling: }\quad This upsamples long documents while ignoring their source domains, and consequently slightly changes the domain mixture.~\citet{togetherllama32k} uses this approach.

\textbf{Upsample Arxiv/ Book/ Github:}\quad This intentionally upsamples Arxiv/Book/Github data, assuming that these domains are more likely to contain naturally-occurring long documents. For example, YaRN Mistral~\citep{peng2023yarn} and MPT-storyteller~\citep{MosaicML2023Introducing} upsamples books. This approach changes both the domain and length distributions.

The above approaches generally cover most data recipes used in prior works discussed in section~\ref{sec:background}.
In our experiments, we will show that:
(1) using the original mixture cutting at 128K is insufficient for precise retrieval over long-context (Fig.~\ref{fig:original_vs_per_source}), and one need to upsample long sequences; 
(2) improved performance in one domain may not transfer and could even hurt another domain (Table~\ref{tab:loss_7b}), showing that  
one needs to balance the mixture ratio of different domains. 

%% file: figures/fig_data_mix.tex
\begin{figure*}[!t]
\small
  \centering
  \includegraphics[width=\linewidth]{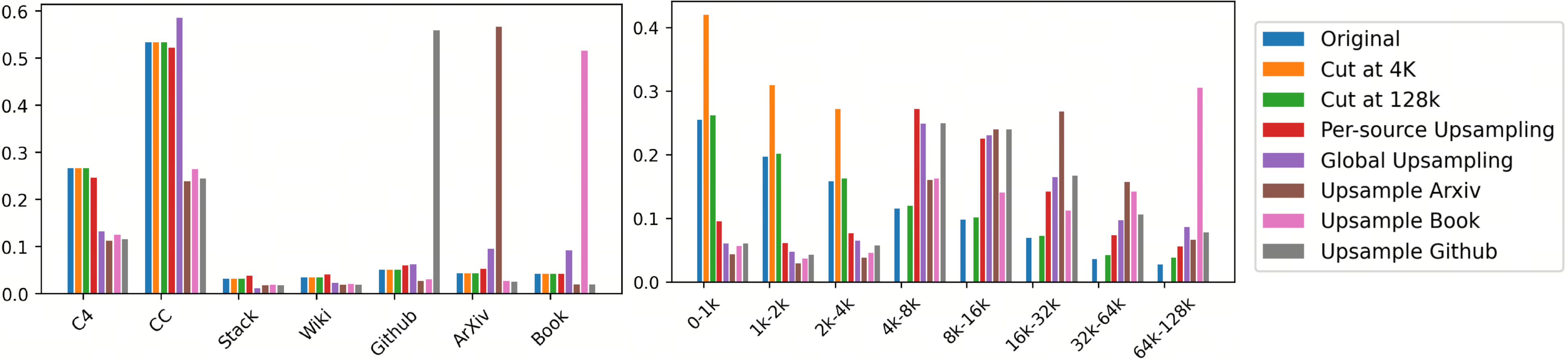}
  \vspace{-6mm}
  \caption{
    Length and domain distributions of the various data mixture strategies. 
    We use  SlimPajama~\citep{cerebras2023slimpajama}, 
    an open-source reproduction of LLaMA~\citep{touvron2023llama} pretraining data mixture,  as the source dataset.
    The \fblue{original data mixture} has about 30\% of documents that are naturally longer than 4K.
    \forange{Cutting documents at 4K}, the common practice of pretraining like~\citet{touvron2023llama}, breaks such long-range dependencies. 
    \fgreen{Cutting documents at 128K} retains the naturally-existing long-range dependencies. 
    \fpurple{Global upsampling} longer sequences slightly changes the data mixture. 
    \fred{Per-source upsampling} longer sequences increases the portion of long sequences while keeping the domain mixture the same. 
    Upsampling \fbrown{Arxiv} / \fcyan{Book} / \fgrey{Github} data simultaneously changes domain and length distributions. 
  }
  \vspace{-2mm}
  \label{fig:data_mix}
\end{figure*}

%% file: 040_exp_setting.tex
We  continue pretraining the model on the data mixture yielded by the "Per-source Upsampling" strategy as introduced in section~\ref{sec:data}, where we train with a 80K sequence length for 7B models, and 64K for 13B models. A ``brute force'' training on 80K context may seem impractical at first glance
given the quadratic complexity of attention.
However, in our initial tests, we found this to be feasible, and the actual wallclock time  was far from quadratic. 
This is due to the fact that most of the time is spent on data transfer from CPU to GPU (since we use offloading), from one GPU to another GPU via NVLink (since we use Zero3, see~\citealp{rajbhandari2020zero}) and from GPU High-Bandwidth Memory (HBM) to Streaming Multiprocessors (SM) (as in FlashAttention, see~\citealp{dao2023flashattention}).  These IO operations are all constant or linear in sequence length.  
The actual quadratic attention computation, when performed on SM, is highly parallelized, and thus largely hidden by the linear IO cost. 
Consequently, training on 80K is only 3x slower than training on 4K. 

Our specific configuration is listed on Table~\ref{tab:infra}.
We note that this configuration is substantially cheaper than previous work~\citep{xiong2023effective}, as they continue pretraining the model on more than 400B tokens. {However this configuration seems to be the limit of Huggingface-DeepSpeed framework, and setting even longer context leads to memory overflow.} 
Since Zero Optimizer is data parallel, increasing number of GPUs cannot further increase context length, and most of the GPU memory is taken by the KV cache.
We refer the reader to  more advanced sequence parallelism techniques in~\citep{li2021sequence, jacobs2023deepspeed} for training on even longer sequences (e.g., 200K), and we leave the exploration of training on 200K context for future work.

For training, we use a constant learning rate 2e-5. 
We modify the base of RoPE positional encoding to adjust it to longer context, as in~\citet{xiong2023effective}.
We pack all data to 80K chunks regardless of the document boundary, following common practice~\citep{raffel2020exploring, touvron2023llama}.
We set the batch size to be 4M tokens.
Note that this batch size is the same as training on 4K context length, as we increase the length of a chunk but decrease the number of chunks in a batch. 
We train the model on 5B tokens, which translates to 5B (size of data) / 4M (batch size) = 2000 optimization steps. 

\input{tables/tab_hardware}

%% file: tables/tab_hardware.tex
\begin{table}[t!]
    \centering
    \caption{We show that long context continual pretraining is {feasible under academic-level resources}.
    Our configuration on 8$\times$80G A100s, as listed below, takes 5 days, which is about 1\% budget than existing works such as~\citet{xiong2023effective}, which trains on 400B tokens.
    }
    \begin{tabular}{@{}ll@{}}
        \toprule
         Framework & Huggingface Transformers  \\ 
          & + DeepSpeed Zero 3  \\
          & + FlashAttention 2  \\
          & + Gradient Checkpointing  \\
          & + CPU Offloading \\
        \midrule
        Hardware & 8 $\times$ 80G A100 \\
        LLaMA 2 7B      & Ctx. 4K  \quad\quad 3 days / 10B tokens\\
                         & Ctx. 80K \quad 10 days / 10B tokens\\
        LLaMA 2 13B     & Ctx. 4K \quad\quad 5 days / 10B tokens\\
                         & Ctx. 64K \quad 13 days / 10B tokens\\
        \midrule
        Hardware & 2 $\times$ 8 $\times$ 80G A100 \\
        LLaMA 2 7B      & Ctx. 4K  \quad\quad 2 days / 10B tokens\\
                         & Ctx. 80K \quad\; 7 days / 10B tokens\\
        LLaMA 2 13B     & Ctx. 4K  \quad\quad 4 days / 10B tokens\\
                         & Ctx. 64K \quad 10 days / 10B tokens\\
        \bottomrule
    \end{tabular}
    \label{tab:infra}
\end{table}

%% file: 050_exp.tex
We continue pretraining the base LLaMA 2 7B/ 13B models on data produced by our "Per-source Upsampling" strategy (Sec.~\ref{ssec:data_mixture}).
The number of training tokens is 5B. 
We first show that our method outperforms strong open-source baselines like YaRN Mistral 7B 128K~\citep{peng2023yarn} and LongLoRA 7B 100K~\citep{chen2023longlora}, and closes the gap to GPT-4 Turbo 128K.
Then we dissect the ingredients of our data recipe. 
In section~\ref{ssec:exp_data_sufficiency}, we discuss data sufficiency (the \underline{5B} part)
and show that as we increase the amount of data from 100M to 1B, the model gradually surfaces the ability to precisely retrieve given information at arbitrary locations within 128K documents. 
In section~\ref{ssec:data_mixture}, we discuss the data mixture (the \underline{per-source length upsampled} part), and show that the original mixture of SlimPajama is insufficient (although it has about 30\% data longer than 4K), and one needs to upsample long sequences for the model to acquire the long context capability. 
We also show that the improvements on one upsampled domain (e.g., Books), have limited transfer to others, and can even hurt them.
The Book and Code domains are a particular example that improvements on one come with detrimental impact on the other. 

Existing work heavily uses validation loss as the primary evaluation \citep{chen2023extending, chen2023longlora, xiao2023efficient, peng2023yarn}. 
We  show that using only the validation loss is insufficient because two data recipes that result in similar loss may have substantially different  retrieval capabilities, as we show in Fig.~\ref{fig:original_vs_per_source}.
In addition to  Needle-in-a-Haystack, we use a real-world book-long question answering benchmark~\citep{zhang2023infinitebench} for our evaluation.
There are earlier long context benchmarks such as Zeroscrolls~\citep{shaham2023zeroscrolls}, longbench~\citep{bai2023longbench} and L-Eval~\citep{an2023eval}.
We do not evaluate on them because their lengths are mostly around 10K, which were considered long at their release time, but substantially shorter than the 128K regime, which is the focus of the present work.

\subsection{Overall Performance}
\label{ssec:exp_overall}

\input{051_overall}

\subsection{Data Quantity}
\label{ssec:exp_data_sufficiency}
\input{051_data_sufficiency}

\subsection{Data Mixture}
\label{ssec:data_mixture}
\input{051_data_mixture}

%% file: 051_overall.tex
\input{tables/tab_benchmark}
Figure~\ref{fig:needle_in_a_haystack} compares our method with Together AI's LLaMA-2 32K~\citep{togetherllama32k}, LMSys' LongChat v1.5 32K~\citep{longchat2023}, YaRN Mistral 7B 128K~\citep{peng2023yarn}, and LongLoRA 100K~\citep{chen2023longlora}.
We choose these models as our baseline because they are so far the top open-source long-context language models (as evidenced by thousands of GitHub stars). 
LongChat v1.5 32K does not perform well even for sequence length 16K-32K, and we hypothesize that this is because the model has not gone through enough continual pretraining, but directly goes to finetuning. 
Together AI's LLaMA-2 7B's data mixture is similar to ours, which we believe to be the reason that they perform well within 32K. 
Yet their model does not generalize beyond 32K, while ours generalizes from 80K to 128K. 
For LongLoRA, we believe the two major problems are that they use sparse attention and that they do not upsample long sequence data, which we show is  problematic in Fig.~\ref{fig:original_vs_per_source}.
For YaRN Mistral, we hypothesize that its underperformance is due to its being only trained on PG19 book~\citep{raecompressive2019}; we show that it is important to use a balanced domain mixture in Table~\ref{tab:loss_7b}.

\input{figures/fig_data_scaling_needle}

In Table~\ref{tab:benchmark} we show that our method not only improves precise retrieval, but maintains short context performance, evidenced by strong MMLU~\citep{hendrycks2020measuring} score, which is a widely-accepted benchmark for   testing the general capabilities (within short context) of language models, and is used by Chinchilla~\citep{hoffmann2022training}, LLama~\citep{touvron2023llama, touvron2023llama2}, Mistral~\citep{jiang2023mistral}, QWen~\citep{bai2023qwen}, and Baichuan~\citep{yang2023baichuan}. 

Table~\ref{tab:long_qa} further compares 128K context language models on a book-long question answering benchmark recently proposed by~\citet{zhang2023infinitebench}. 
This task conditions language models on a book and then asks questions about the plot. 
Our method outperforms LongLoRA and Yarn Mistral (even though Mistral 7B is a stronger base model than LLaMA 2 7B we use). 
Our 13B model performance closes the gap to GPT-4 128K, and we anticipate that future scaling and instruction tuning will further improve  performance. 
While there are other long-context benchmarks in InfiniBench~\citep{zhang2023infinitebench},  in our initial experiments we found that  models often had trouble understanding the instruction (because they are not instruction tuned). Hence we focus on the BookQA benchmark where base LLMs performed reasonably without instruction tuning.

%% file: tables/tab_benchmark.tex
\begin{table}[t!]
\centering
\caption{
Our method achieves better long context task performance  (Needle. performance) \textit{without}  compromising short context performance (MMLU).
}
\begin{tabular}{@{}ll|cc@{}}
\toprule
                   & \bf Ctx.  & \bf Needle.   & \bf MMLU   \\ 
\midrule  
\multicolumn{4}{@{}l}{\bf Non-LLaMA Models}\\
GPT-4-Turbo \quad &128K          & \bf 87.1  &\bf 86.4     \\
GPT-3.5-Turbo \quad &16K         & -   &67.3     \\
YaRN Mistral 7B &128K            & 57.4  & 59.4  \\ 
\midrule
\multicolumn{4}{@{}l}{\bf LLaMA-2 7B Based Models} \\
Together LLaMA-2 7B & 32K     & 27.9  & \bf 44.8    \\ 
LongChat v1.5 7B  & 32K       & 18.0   & 42.3    \\ 
LongLoRA 7B       & 100K      & 70.0  & 37.9    \\ 
Ours LLaMA-2 7B   & 80K       & \bf 88.0   & 43.3    \\
\midrule
\multicolumn{4}{@{}l}{\bf LLaMA-2 13B Based Models} \\
LongLoRA 13B      & 64K       & 54.1  & 50.1    \\ 
Ours LLaMA-2 13B  & 64K       &\bf 90.0  &\bf 52.4    \\
\bottomrule
\end{tabular}
\label{tab:benchmark}
\end{table}


%% file: figures/fig_data_scaling_needle.tex
\begin{figure*}[!t]
\small
  \centering
  \includegraphics[width=\linewidth]{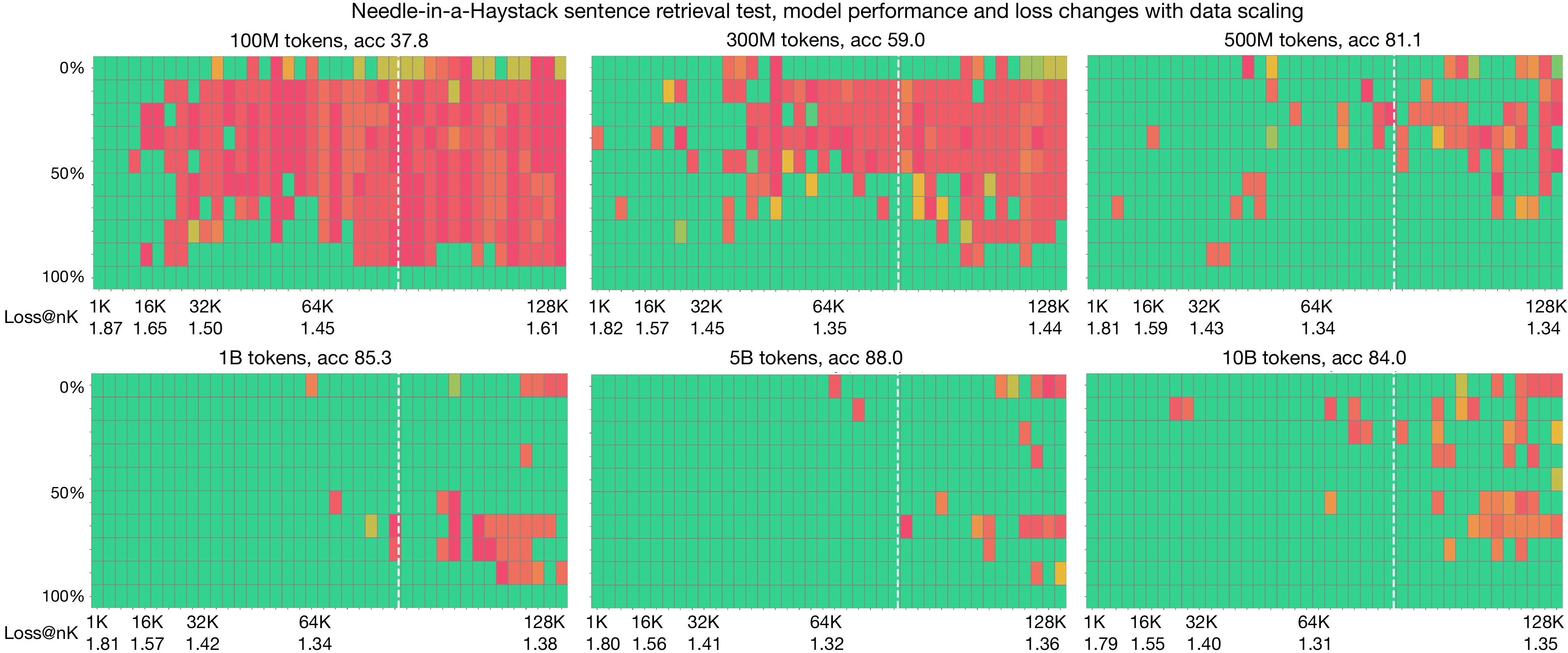}
  \vspace{-5mm}
  \caption{
    As we increase the number of trained tokens, the model's retrieval performance increases with converged validation loss. 
    We further note two observations with regard to data quantity: 
    (1)    500M tokens are enough to unlock  most of the retrieval accuracy. Since we are not explicitly training the model to perform retrieval, this potentially suggests that  the model already has the precise retrieval capability, and that we are simply extending this capability with lightweight continual pretraining;
    (2) the model's retrieval performance saturates at about 5B tokens, and further scaling to 10B tokens does not improve length generalization.
  }
  \vspace{-2mm}
  \label{fig:needle_data_scaling}
\end{figure*}

%% file: 051_data_sufficiency.tex
\input{tables/tab_long_qa}

\input{tables/tab_loss_7b}

Now we consider data quantity, i.e., how much data enables the language model to utilize the information at arbitrary locations within 128K context.
Our hypothesis is that precise retreival over long-range context is an intrinsic capability obtained by large-scale pretraining, even when the pretraining context length is substantially shorter (4K in many cases). 
If this hypothesis is true, then lightweight continual pretraining should be enough to extend this capability to much longer context lengths than see in training. That is, we would not need data-intensive continual pretraining as used by ~\citet{xiong2023effective} and \citet{xverse}.

Fig.~\ref{fig:needle_data_scaling} shows how data scaling gradually unlocks the model's retrieval capability. 
When the model is continually pretrained on 100M to 300M tokens, the model's loss gradually decreases, but it has not converged yet. 
At 500M to 1B tokens, the model achieves relatively good performance within its continually pretrained 80K context, but does not generalize to 80K-128K range. 
After 5B tokens, the model performs well on 0-80K, and can generalize to unseen lengths 80K-128K. 
At 10B token, the model seems to overfit on its 80K training range, and the length generalization starts to decrease. 
Along this process, we see a gradual decrease in loss, which   correlates with the increase in the retrieval performance. 
This result provides more evidence for our hypothesis, as only 500M tokens are required to enable a model's retrieval capability with 80K context. 

Continually pretraining a 7B LLaMA-2 takes about 5 days on 8$\times$80G A100 GPUs, and the cost is reasonable compared to large-scale pretraining (which requires months of compute on thousands of GPUs). Our results suggest that for supervised finetuning, since training on long-context is substantially cheaper than previously thought, future work may dive deeper on the solutions for 100K length finetuning and reasoning, which so far has almost no open-source work to our knowledge. 
For pretraining research, currently there is no definite answer as to whether long-context continual pretraining should be combined with other capabilities, such as math~\citep{azerbayev2023llemma} and code~\citep{chen2021evaluating}, which typically require hundreds of billions of tokens. Our results suggest that long-context continual pretraining could be a separate stage after code and math  pretraining.

%% file: tables/tab_long_qa.tex
\begin{table}[t!]
\centering
\vspace{-2mm}
\caption{
The performaince gain shown in Needle-in-a-Haystack further translates to improvements on downstream BookQA.
We report the results on  128K length InfiniBench long book question answering task~\citep{zhang2023infinitebench}. 
Our method outperforms top open-source models and closes the gap to GPT-4. 
}
\begin{tabular}{@{}lcc@{}}
\toprule
\bf Model / Train Len& \bf Test Len  & \bf BookQA      \\ 
\midrule  
GPT-4-Turbo 128K     & 128K     & \bf 37.4 \\
LongLoRA 7B 100K     & 128K     & 24.3  \\ 
Ours LLaMA-2 7B 80K  & 128K     & 27.4  \\
LongLoRA 13B 64K     & 128K     & 24.6  \\ 
YaRN Mistral 7B 128K & 128K     & 26.3  \\ 
Ours LLaMA-2 13B 64K & 128K     & \bf 31.1  \\
\bottomrule
\end{tabular}
\vspace{-4mm}
\label{tab:long_qa}
\end{table}

%% file: tables/tab_loss_7b.tex
\begin{table*}[t!]
\centering
\caption{
How different data engineering methods improves (or does not improve) the original data mixture, where we show results from continual pretraining of a 7B LLaMA-2 on 5B tokens packed to 80K length.
We consider how changing the original data mixture to long-context upsampled data mixture results in different loss across domains, and report the \textit{loss differences}  to the original data mixture (e.g., the loss of per-source length upsampling minus the loss of original mixture).
We view a loss difference larger than 0.01 significant, 
marked by a \redb{red} shade indicating performance decrease of larger than \redb{+0.01} loss; or a \greenb{green} shade indicating performance improvements of smaller than \greenb{-0.01} loss, or a \greyb{grey}  shade indicating no significant differences.
Although upsampling book / code / arxiv improves in-domain performance for both short and long context, such improvements do {not} transfer to all domains. Instead, upsampling one domain, e.g., code, may even harm another domain, e.g., book. 
\textit{Per-source length upsampling is the most balanced mixture} with almost no significant increase of loss across domains.
}
\begin{tabular}{@{}llllllll@{}}
\toprule
 & \bf C4               & \bf CC              & \bf Stack       & \bf Arxiv           & \bf Wiki            & \bf Book            & \bf Github  \\ 
\midrule 
\multicolumn{8}{@{}l@{}}{\bf 0 - 4K Context Length} \\ 
\cmidrule{1-2}
Original             & 2.038               & 1.760              & 1.519              & 1.660                & 1.424         & 2.085         & 0.907         \\
v.s. Per-source      & \greyb{\small + .002}& \greyb{\small+ .008}& \greyb{\small-- .001}& \greyb{\small-- .008}  & \greenb{\small-- .040} & \greenb{\small-- .065} & \greyb{\small-- .008} \\
v.s. Global          & \greyb{\small + .008}& \lredb{\small+ .010} & \lredb{\small+ .015} & \greenb{\small-- .020} & \greenb{\small-- .020} & \greenb{\small-- .140} & \lredb{\small+ .015} \\
v.s. Code$\uparrow$  & \lredb{\small + .010} & \lredb{\small+ .016} & \lredb{\small+ .010} & \greyb{\small+ .006}  & \greenb{\small-- .026} & \redb{\small+ .030}   & \greenb{\small-- .023} \\
v.s. Book$\uparrow$  & \lredb{\small + .010} & \lredb{\small+ .016} & \redb{\small+ .021} & \greyb{\small+ .000}  & \lgreenb{\small-- .010} & \greenb{\small-- .175} & \redb{\small+ .029} \\
v.s. Arxiv$\uparrow$ & \greyb{\small + .006} & \lredb{\small+ .016} & \lredb{\small+ .013} & \greenb{\small-- .060} & \greenb{\small-- .030} & \redb{\small+ .040}   & \redb{\small+ .025}\\
\midrule
\multicolumn{8}{@{}l@{}}{\bf 4K - 128K Context Length} \\ 
\cmidrule{1-2}
Original             & 1.560                 & 1.650                & 0.786              & 1.075                & 1.313         & 1.852         & 0.447         \\
v.s. Per-source      & \lgreenb{\small-- .010}  &\lgreenb{\small-- .010}  &\greyb{\small-- .006}  &\lgreenb{\small-- .011}  &\greenb{\small-- .044}  &\lgreenb{\small-- .014}  &\greyb{\small+ .002}   \\
v.s. Global          & \lgreenb{\small-- .010}  &\greyb{\small-- .006}  &\greyb{\small-- .001}  &\lgreenb{\small-- .016}  &\greenb{\small-- .040}  &\lgreenb{\small-- .018}  &\greyb{\small-- .007}   \\
v.s. Code$\uparrow$  & \greyb{\small-- .008}  &\greyb{\small-- .002}  &\greyb{\small-- .003}  &\greyb{\small-- .007}  &\greenb{\small-- .042}  &\greyb{\small-- .010}  &\greenb{\small-- .029} \\
v.s. Book$\uparrow$  & \lgreenb{\small-- .010}  &\greyb{\small-- .006}  &\greyb{\small+ .001}  &\greyb{\small-- .007}  &\greenb{\small-- .037}  &\greenb{\small-- 0.30}  &\greyb{\small+ .000}  \\
v.s. Arxiv$\uparrow$ & \greyb{\small-- .008}  &\greyb{\small-- .002}  &\greyb{\small+ .002}  &\greenb{\small-- .036}  &\greenb{\small-- .039}  &\greyb{\small-- .010}  &\greyb{\small-- .004}  \\
\bottomrule
\end{tabular}
\label{tab:loss_7b}
\end{table*}

%% file: 051_data_mixture.tex
\input{figures/fig_original_vs_per_source}

Now we discuss why \textit{per-source length upsampling} is necessary when constructing the data mixture. 
Recall that this strategy keeps the mixture ratio of the data sources the same as the original data, i.e., 67\% CommonCrawl (CC), 15\% C4, 4.5\% Github, 4.5\% Wikipedia, 4.5\% books, 2.5\% Arxiv and  2.0\% StackExchange for SlimPajama. 
Then in each of the domains, we upsample sequences longer than 4K from about 30\% to about 70\%.
Doing this enables us to keep the domain mixture ratio fixed, only changing the length distribution of the training documents. 
In contrast, globally upsampling long sequences (without considering their domain), or intentionally upsampling code/ book/ Arxiv (since they are long) changes both the domain mixture and the length distribution. 

We compare all the data mixtures with the original one packed to 80K / 64K for the continual pretraining of the 7B / 13B LLaMA-2. 
This baseline approach effectively never changes the original data, but keeps the naturally existing long sequence that would otherwise been cut to 4K (recall Sec.~\ref{sec:data}). 
If a data mixture is effective, it should be at least better than this baseline ``do nothing'' approach.

Table~\ref{tab:loss_7b} compares the per-domain loss differences of all the data mixture against the baseline original mixture. 
We report the differences of the validation loss, where a more than 0.01 loss change is considered significant, following common pretraining practice~\citep{kaplan2020scaling, peng2023yarn, hoffmann2022training}. 
We observe: 
(1) for 0-4K short context data, most of the data mixtures have a negative impact on webpages (C4, CC and StackExchange), except for the per-source approach;
(2) performance improvements from domains like Book may not transfer to others, and even hurt code performance;
(3) per-source length upsampling is the most balanced mixture that improves 4K-128K context losses without much sacrificing short-context losses, whereas all other methods show more or less performance tradeoff across domains. 
Combining these observations, we recommend the per-source length-upsampling strategy. 

Since per-source length upsampling keeps the domain mixture ratio while increasing tokens from long sequences, its major difference to the original data mixture is the distribution of sequence length. 
Fig.~\ref{fig:original_vs_per_source} makes a close comparison between the two data strategies. 
Specifically, we train two LLaMA-2 7B models with a 80K context length using original / per-source upsampled data mixture, and report their per-length validation loss and Needle-in-a-Haystack performance.
Note that LongLoRA~\citep{chen2023longlora} uses the original data mixture without length upsampling, 
so our results also explains why we achieve better performance than LongLoRA (Fig.~\ref{fig:needle_in_a_haystack}). 
We see that the original data mixture without length upsampling, despite achieving a very close loss, underperforms on precise retrieval. 
Per-source length upsampling significantly improves precise retrieval. 
This observation also serves as strong evidence why only using test loss, the evaluation used in most prior work \cite{chen2023extending, peng2023yarn, chen2023longlora, xiao2023efficient, claude}, may conceal the underlying model differences.
The necessity of length upsampling is demonstrated by the Needle-in-a-Haystack test.

%% file: figures/fig_original_vs_per_source.tex
\begin{figure}[!t]
\small
  \centering
  \includegraphics[width=\linewidth]{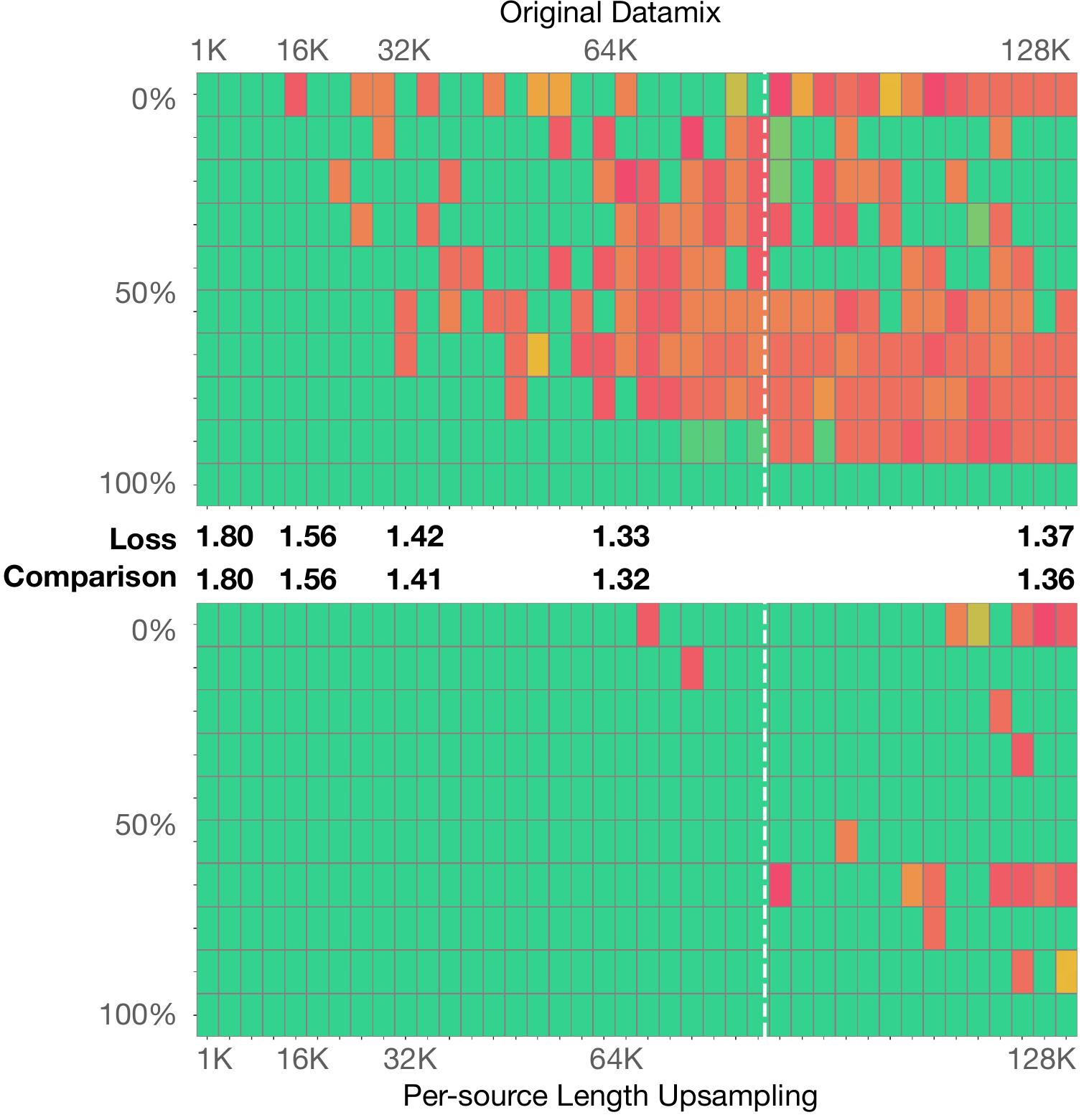}
    \vspace{-6mm}
  \caption{The importance of length upsampling.
    The validation loss, 
    a common measure for long-context language modeling used in~\citet{xiong2023effective, chen2023longlora}, 
    {cannot} reveal long-context retreival capability, but Needle-in-a-Haystack can. 
    The original data mixture, i.e. without length upsampling, despite giving very close loss, performs badly on precise retrieval. Per-source length upsampling significantly improves precise retrieval. 
  }
  \vspace{-4mm}
  \label{fig:original_vs_per_source}
\end{figure}

%% file: 060_discussion.tex
We attribute our improvements over strong open-source baselines, as is detailed in section~\ref{sec:exp},
to our careful treatments of data engineering.
Our results echo  the recent wisdom that in large language model research, 
data engineering is equally important as modeling and algorithmic innovations~\citep{kaplan2020scaling, hoffmann2022training, brown2020language}.
As we list in  Table~\ref{tab:design_diff}, many of the data details are crucial for strong long-context performance, yet may be easily overlooked. We also acknowledge that our research is made possible through utilizing the latest innovations in machine learning systems research, particularly FlashAttention~\citep{dao2023flashattention} as it reduces the memory usage of the attention mechanism from quadratic to linear. Incorporating further work on sequence parallelism~\citep{li2021sequence, jacobs2023deepspeed}  could enable brute force training of even longer context (e.g., 200K) models.

The Transformer's use of position embeddings makes it difficult to generalize significantly beyond contexts seen during training even with relative positional encodings~\citep{alibi2021,xpos2022,fire2024}, thus necessitating (at least a little bit of) continual pretraining. There has much recent work on RNN-like architectures, which implicitly encode positional information through their hidden states, that perform competitively with Transformers \cite{sun2023retentive,HGRN,Gu2023MambaLS,yang2023gla}. It would be interesting to test whether such models can generalize zero-shot to longer contexts than seen during training  on the Needle-in-a-Haystack benchmark.

Long-context language model research at the 100K-level is still a developing research area.
This work only studies continual pretraining, and  research on instruction finetuning language models on tasks of 100K context length (e.g., repo-level code understanding) is still limited. 
So far there seems to no open-source instruction-finetuned 100K context language models. We hope our work serve as a basis for future work on 100K-level long context superivsed finetuning.

%% file: 070_conclusion.tex
This work studies a continual pretraining recipe for scaling language models' context length to 128K tokens. 
We demonstrate that the ability to utilize information at arbitrary locations within the 128K input is  already mostly acquired by large-scale pretraining, even for models pretrained on substantially shorter 4K context. 
We show that continual pretraining of the full model on 1-5 billion tokens from per-source length-upsampled data gives the most balanced performance improvements. Our work closes the gap to frontier models like GPT-4 128K on the retrieval task and serves as the foundation for future long-context instruction finetuning research.

%% file: main.bbl
\begin{thebibliography}{45}
\providecommand{\natexlab}[1]{#1}
\providecommand{\url}[1]{\texttt{#1}}
\expandafter\ifx\csname urlstyle\endcsname\relax
  \providecommand{\doi}[1]{doi: #1}\else
  \providecommand{\doi}{doi: \begingroup \urlstyle{rm}\Url}\fi

\bibitem[An et~al.(2023)An, Gong, Zhong, Li, Zhang, Kong, and Qiu]{an2023eval}
An, C., Gong, S., Zhong, M., Li, M., Zhang, J., Kong, L., and Qiu, X.
\newblock L-eval: Instituting standardized evaluation for long context language
  models.
\newblock \emph{arXiv preprint arXiv:2307.11088}, 2023.

\bibitem[Anthropic(2023)]{claude}
Anthropic.
\newblock Model card and evaluations for claude models, July 2023.
\newblock URL \url{https://www.anthropic.com/product}.

\bibitem[Azerbayev et~al.(2023)Azerbayev, Schoelkopf, Paster, Santos, McAleer,
  Jiang, Deng, Biderman, and Welleck]{azerbayev2023llemma}
Azerbayev, Z., Schoelkopf, H., Paster, K., Santos, M.~D., McAleer, S., Jiang,
  A.~Q., Deng, J., Biderman, S., and Welleck, S.
\newblock Llemma: An open language model for mathematics.
\newblock \emph{arXiv preprint arXiv:2310.10631}, 2023.

\bibitem[Bai et~al.(2023{\natexlab{a}})Bai, Bai, Chu, Cui, Dang, Deng, Fan, Ge,
  Han, Huang, et~al.]{bai2023qwen}
Bai, J., Bai, S., Chu, Y., Cui, Z., Dang, K., Deng, X., Fan, Y., Ge, W., Han,
  Y., Huang, F., et~al.
\newblock Qwen technical report.
\newblock \emph{arXiv preprint arXiv:2309.16609}, 2023{\natexlab{a}}.

\bibitem[Bai et~al.(2023{\natexlab{b}})Bai, Lv, Zhang, Lyu, Tang, Huang, Du,
  Liu, Zeng, Hou, et~al.]{bai2023longbench}
Bai, Y., Lv, X., Zhang, J., Lyu, H., Tang, J., Huang, Z., Du, Z., Liu, X.,
  Zeng, A., Hou, L., et~al.
\newblock Longbench: A bilingual, multitask benchmark for long context
  understanding.
\newblock \emph{arXiv preprint arXiv:2308.14508}, 2023{\natexlab{b}}.

\bibitem[Bairi et~al.(2023)Bairi, Sonwane, Kanade, Iyer, Parthasarathy,
  Rajamani, Ashok, Shet, et~al.]{bairi2023codeplan}
Bairi, R., Sonwane, A., Kanade, A., Iyer, A., Parthasarathy, S., Rajamani, S.,
  Ashok, B., Shet, S., et~al.
\newblock Codeplan: Repository-level coding using llms and planning.
\newblock \emph{arXiv preprint arXiv:2309.12499}, 2023.

\bibitem[Brown et~al.(2020)Brown, Mann, Ryder, Subbiah, Kaplan, Dhariwal,
  Neelakantan, Shyam, Sastry, Askell, et~al.]{brown2020language}
Brown, T., Mann, B., Ryder, N., Subbiah, M., Kaplan, J.~D., Dhariwal, P.,
  Neelakantan, A., Shyam, P., Sastry, G., Askell, A., et~al.
\newblock Language models are few-shot learners.
\newblock \emph{Advances in neural information processing systems},
  33:\penalty0 1877--1901, 2020.

\bibitem[Caciularu et~al.(2023)Caciularu, Peters, Goldberger, Dagan, and
  Cohan]{caciularu2023peek}
Caciularu, A., Peters, M.~E., Goldberger, J., Dagan, I., and Cohan, A.
\newblock Peek across: Improving multi-document modeling via cross-document
  question-answering.
\newblock \emph{arXiv preprint arXiv:2305.15387}, 2023.

\bibitem[Chen et~al.(2021)Chen, Tworek, Jun, Yuan, Pinto, Kaplan, Edwards,
  Burda, Joseph, Brockman, et~al.]{chen2021evaluating}
Chen, M., Tworek, J., Jun, H., Yuan, Q., Pinto, H. P. d.~O., Kaplan, J.,
  Edwards, H., Burda, Y., Joseph, N., Brockman, G., et~al.
\newblock Evaluating large language models trained on code.
\newblock \emph{arXiv preprint arXiv:2107.03374}, 2021.

\bibitem[Chen et~al.(2023{\natexlab{a}})Chen, Wong, Chen, and
  Tian]{chen2023extending}
Chen, S., Wong, S., Chen, L., and Tian, Y.
\newblock Extending context window of large language models via positional
  interpolation.
\newblock \emph{arXiv preprint arXiv:2306.15595}, 2023{\natexlab{a}}.

\bibitem[Chen et~al.(2023{\natexlab{b}})Chen, Qian, Tang, Lai, Liu, Han, and
  Jia]{chen2023longlora}
Chen, Y., Qian, S., Tang, H., Lai, X., Liu, Z., Han, S., and Jia, J.
\newblock Longlora: Efficient fine-tuning of long-context large language
  models.
\newblock \emph{arXiv preprint arXiv:2309.12307}, 2023{\natexlab{b}}.

\bibitem[Dao(2023)]{dao2023flashattention}
Dao, T.
\newblock Flashattention-2: Faster attention with better parallelism and work
  partitioning.
\newblock \emph{arXiv preprint arXiv:2307.08691}, 2023.

\bibitem[Fuzhao~Xue \& You(2023)Fuzhao~Xue and You]{openmoe2023}
Fuzhao~Xue, Zian~Zheng, Y. F. J. N. Z. Z. W.~Z. and You, Y.
\newblock Openmoe: Open mixture-of-experts language models.
\newblock \url{https://github.com/XueFuzhao/OpenMoE}, 2023.

\bibitem[Gu \& Dao(2023)Gu and Dao]{Gu2023MambaLS}
Gu, A. and Dao, T.
\newblock Mamba: Linear-time sequence modeling with selective state spaces.
\newblock 2023.
\newblock URL \url{https://api.semanticscholar.org/CorpusID:265551773}.

\bibitem[Hendrycks et~al.(2020)Hendrycks, Burns, Basart, Zou, Mazeika, Song,
  and Steinhardt]{hendrycks2020measuring}
Hendrycks, D., Burns, C., Basart, S., Zou, A., Mazeika, M., Song, D., and
  Steinhardt, J.
\newblock Measuring massive multitask language understanding.
\newblock \emph{arXiv preprint arXiv:2009.03300}, 2020.

\bibitem[Hoffmann et~al.(2022)Hoffmann, Borgeaud, Mensch, Buchatskaya, Cai,
  Rutherford, Casas, Hendricks, Welbl, Clark, et~al.]{hoffmann2022training}
Hoffmann, J., Borgeaud, S., Mensch, A., Buchatskaya, E., Cai, T., Rutherford,
  E., Casas, D. d.~L., Hendricks, L.~A., Welbl, J., Clark, A., et~al.
\newblock Training compute-optimal large language models.
\newblock \emph{arXiv preprint arXiv:2203.15556}, 2022.

\bibitem[Jacobs et~al.(2023)Jacobs, Tanaka, Zhang, Zhang, Song, Rajbhandari,
  and He]{jacobs2023deepspeed}
Jacobs, S.~A., Tanaka, M., Zhang, C., Zhang, M., Song, L., Rajbhandari, S., and
  He, Y.
\newblock Deepspeed ulysses: System optimizations for enabling training of
  extreme long sequence transformer models.
\newblock \emph{arXiv preprint arXiv:2309.14509}, 2023.

\bibitem[Jiang et~al.(2023)Jiang, Sablayrolles, Mensch, Bamford, Chaplot,
  Casas, Bressand, Lengyel, Lample, Saulnier, et~al.]{jiang2023mistral}
Jiang, A.~Q., Sablayrolles, A., Mensch, A., Bamford, C., Chaplot, D.~S., Casas,
  D. d.~l., Bressand, F., Lengyel, G., Lample, G., Saulnier, L., et~al.
\newblock Mistral 7b.
\newblock \emph{arXiv preprint arXiv:2310.06825}, 2023.

\bibitem[Kamradt(2023)]{needleinhaystack}
Kamradt, G.
\newblock Needle in a haystack - pressure testing llms.
\newblock \url{https://github.com/gkamradt/LLMTest_NeedleInAHaystack}, 2023.

\bibitem[Kaplan et~al.(2020)Kaplan, McCandlish, Henighan, Brown, Chess, Child,
  Gray, Radford, Wu, and Amodei]{kaplan2020scaling}
Kaplan, J., McCandlish, S., Henighan, T., Brown, T.~B., Chess, B., Child, R.,
  Gray, S., Radford, A., Wu, J., and Amodei, D.
\newblock Scaling laws for neural language models.
\newblock \emph{arXiv preprint arXiv:2001.08361}, 2020.

\bibitem[Li et~al.(2023{\natexlab{a}})Li, Shao, Xie, Sheng, Zheng, Gonzalez,
  Stoica, Ma, and Zhang]{longchat2023}
Li, D., Shao, R., Xie, A., Sheng, Y., Zheng, L., Gonzalez, J.~E., Stoica, I.,
  Ma, X., and Zhang, H.
\newblock How long can open-source llms truly promise on context length?, June
  2023{\natexlab{a}}.
\newblock URL \url{https://lmsys.org/blog/2023-06-29-longchat}.

\bibitem[Li et~al.(2021)Li, Xue, Baranwal, Li, and You]{li2021sequence}
Li, S., Xue, F., Baranwal, C., Li, Y., and You, Y.
\newblock Sequence parallelism: Long sequence training from system perspective.
\newblock \emph{arXiv preprint arXiv:2105.13120}, 2021.

\bibitem[Li et~al.(2023{\natexlab{b}})Li, You, Guruganesh, Ainslie, Ontanon,
  Zaheer, Sanghai, Yang, Kumar, and Bhojanapalli]{fire2024}
Li, S., You, C., Guruganesh, G., Ainslie, J., Ontanon, S., Zaheer, M., Sanghai,
  S., Yang, Y., Kumar, S., and Bhojanapalli, S.
\newblock Functional interpolation for relative positions improves long context
  transformers.
\newblock \emph{arXiv preprint arXiv:2310.04418}, 2023{\natexlab{b}}.

\bibitem[Mazumder \& Liu(2024)Mazumder and Liu]{mazumder2024lifelong}
Mazumder, S. and Liu, B.
\newblock \emph{Lifelong and Continual Learning Dialogue Systems}.
\newblock Springer Nature, 2024.

\bibitem[Peng et~al.(2023)Peng, Quesnelle, Fan, and Shippole]{peng2023yarn}
Peng, B., Quesnelle, J., Fan, H., and Shippole, E.
\newblock Yarn: Efficient context window extension of large language models,
  2023.

\bibitem[Press et~al.(2021)Press, Smith, and Lewis]{alibi2021}
Press, O., Smith, N.~A., and Lewis, M.
\newblock Train short, test long: Attention with linear biases enables input
  length extrapolation.
\newblock \emph{arXiv preprint arXiv:2108.12409}, 2021.

\bibitem[Qin et~al.(2023)Qin, Yang, and Zhong]{HGRN}
Qin, Z., Yang, S., and Zhong, Y.
\newblock Hierarchically gated recurrent neural network for sequence modeling.
\newblock \emph{CoRR}, abs/2311.04823, 2023.
\newblock \doi{10.48550/ARXIV.2311.04823}.
\newblock URL \url{https://doi.org/10.48550/arXiv.2311.04823}.

\bibitem[Rae et~al.(2019)Rae, Potapenko, Jayakumar, Hillier, and
  Lillicrap]{raecompressive2019}
Rae, J.~W., Potapenko, A., Jayakumar, S.~M., Hillier, C., and Lillicrap, T.~P.
\newblock Compressive transformers for long-range sequence modelling.
\newblock \emph{arXiv preprint}, 2019.
\newblock URL \url{https://arxiv.org/abs/1911.05507}.

\bibitem[Raffel et~al.(2020)Raffel, Shazeer, Roberts, Lee, Narang, Matena,
  Zhou, Li, and Liu]{raffel2020exploring}
Raffel, C., Shazeer, N., Roberts, A., Lee, K., Narang, S., Matena, M., Zhou,
  Y., Li, W., and Liu, P.~J.
\newblock Exploring the limits of transfer learning with a unified text-to-text
  transformer.
\newblock \emph{The Journal of Machine Learning Research}, 21\penalty0
  (1):\penalty0 5485--5551, 2020.

\bibitem[Rajbhandari et~al.(2020)Rajbhandari, Rasley, Ruwase, and
  He]{rajbhandari2020zero}
Rajbhandari, S., Rasley, J., Ruwase, O., and He, Y.
\newblock Zero: Memory optimizations toward training trillion parameter models.
\newblock In \emph{SC20: International Conference for High Performance
  Computing, Networking, Storage and Analysis}, pp.\  1--16. IEEE, 2020.

\bibitem[Shaham et~al.(2023)Shaham, Ivgi, Efrat, Berant, and
  Levy]{shaham2023zeroscrolls}
Shaham, U., Ivgi, M., Efrat, A., Berant, J., and Levy, O.
\newblock Zeroscrolls: A zero-shot benchmark for long text understanding, 2023.

\bibitem[Soboleva et~al.(2023)Soboleva, Al-Khateeb, Myers, Steeves, Hestness,
  and Dey]{cerebras2023slimpajama}
Soboleva, D., Al-Khateeb, F., Myers, R., Steeves, J.~R., Hestness, J., and Dey,
  N.
\newblock {SlimPajama: A 627B token cleaned and deduplicated version of
  RedPajama}.
\newblock
  \url{https://www.cerebras.net/blog/slimpajama-a-627b-token-cleaned-and-deduplicated-version-of-redpajama},
  2023.
\newblock URL \url{https://huggingface.co/datasets/cerebras/SlimPajama-627B}.

\bibitem[Sun et~al.(2022)Sun, Dong, Patra, Ma, Huang, Benhaim, Chaudhary, Song,
  and Wei]{xpos2022}
Sun, Y., Dong, L., Patra, B., Ma, S., Huang, S., Benhaim, A., Chaudhary, V.,
  Song, X., and Wei, F.
\newblock A length-extrapolatable transformer.
\newblock \emph{arXiv preprint arXiv:2212.10554}, 2022.

\bibitem[Sun et~al.(2023)Sun, Dong, Huang, Ma, Xia, Xue, Wang, and
  Wei]{sun2023retentive}
Sun, Y., Dong, L., Huang, S., Ma, S., Xia, Y., Xue, J., Wang, J., and Wei, F.
\newblock Retentive network: A successor to transformer for large language
  models.
\newblock \emph{arXiv preprint arXiv:2307.08621}, 2023.

\bibitem[Team(2023)]{MosaicML2023Introducing}
Team, M.~N.
\newblock Introducing mpt-7b: A new standard for open-source, commercially
  usable llms, 2023.
\newblock URL \url{www.mosaicml.com/blog/mpt-7b}.
\newblock Accessed: 2023-03-28.

\bibitem[Together(2023)]{togetherllama32k}
Together.
\newblock Llama-2-7b-32k-instruct — and fine-tuning for llama-2 models with
  together api, August 2023.
\newblock URL \url{https://www.together.ai/blog/llama-2-7b-32k-instruct}.

\bibitem[Touvron et~al.(2023{\natexlab{a}})Touvron, Lavril, Izacard, Martinet,
  Lachaux, Lacroix, Rozi{\`e}re, Goyal, Hambro, Azhar,
  et~al.]{touvron2023llama}
Touvron, H., Lavril, T., Izacard, G., Martinet, X., Lachaux, M.-A., Lacroix,
  T., Rozi{\`e}re, B., Goyal, N., Hambro, E., Azhar, F., et~al.
\newblock Llama: Open and efficient foundation language models.
\newblock \emph{arXiv preprint arXiv:2302.13971}, 2023{\natexlab{a}}.

\bibitem[Touvron et~al.(2023{\natexlab{b}})Touvron, Martin, Stone, Albert,
  Almahairi, Babaei, Bashlykov, Batra, Bhargava, Bhosale,
  et~al.]{touvron2023llama2}
Touvron, H., Martin, L., Stone, K., Albert, P., Almahairi, A., Babaei, Y.,
  Bashlykov, N., Batra, S., Bhargava, P., Bhosale, S., et~al.
\newblock Llama 2: Open foundation and fine-tuned chat models.
\newblock \emph{arXiv preprint arXiv:2307.09288}, 2023{\natexlab{b}}.

\bibitem[Weng(2023)]{weng2023prompt}
Weng, L.
\newblock Llm-powered autonomous agents.
\newblock \emph{lilianweng.github.io}, Jun 2023.
\newblock URL \url{https://lilianweng.github.io/posts/2023-06-23-agent/}.

\bibitem[Xiao et~al.(2023)Xiao, Tian, Chen, Han, and Lewis]{xiao2023efficient}
Xiao, G., Tian, Y., Chen, B., Han, S., and Lewis, M.
\newblock Efficient streaming language models with attention sinks.
\newblock \emph{arXiv preprint arXiv:2309.17453}, 2023.

\bibitem[Xiong et~al.(2023)Xiong, Liu, Molybog, Zhang, Bhargava, Hou, Martin,
  Rungta, Sankararaman, Oguz, et~al.]{xiong2023effective}
Xiong, W., Liu, J., Molybog, I., Zhang, H., Bhargava, P., Hou, R., Martin, L.,
  Rungta, R., Sankararaman, K.~A., Oguz, B., et~al.
\newblock Effective long-context scaling of foundation models.
\newblock \emph{arXiv preprint arXiv:2309.16039}, 2023.

\bibitem[XVerse(2024)]{xverse}
XVerse.
\newblock Xverse-13b, January 2024.
\newblock URL
  \url{https://github.com/xverse-ai/XVERSE-13B/blob/main/README_EN.md}.

\bibitem[Yang et~al.(2023{\natexlab{a}})Yang, Xiao, Wang, Zhang, Bian, Yin, Lv,
  Pan, Wang, Yan, et~al.]{yang2023baichuan}
Yang, A., Xiao, B., Wang, B., Zhang, B., Bian, C., Yin, C., Lv, C., Pan, D.,
  Wang, D., Yan, D., et~al.
\newblock Baichuan 2: Open large-scale language models.
\newblock \emph{arXiv preprint arXiv:2309.10305}, 2023{\natexlab{a}}.

\bibitem[Yang et~al.(2023{\natexlab{b}})Yang, Wang, Shen, Panda, and
  Kim]{yang2023gla}
Yang, S., Wang, B., Shen, Y., Panda, R., and Kim, Y.
\newblock Gated linear attention transformers with hardware-efficient training.
\newblock \emph{arXiv preprint arXiv:2312.06635}, 2023{\natexlab{b}}.

\bibitem[Zhang et~al.(2023)Zhang, Chen, Hu, Wu, Chen, Xu, Dai, Han, Wang, Liu,
  and Sun]{zhang2023infinitebench}
Zhang, X., Chen, Y., Hu, S., Wu, Q., Chen, J., Xu, Z., Dai, Z., Han, X., Wang,
  S., Liu, Z., and Sun, M.
\newblock Infinitebench: 128k long-context benchmark for language models, 2023.

\end{thebibliography}
